\documentclass[hidelinks]{llncs}
\usepackage[utf8]{inputenc}
\usepackage{mathtools}
\usepackage{bbding}
\usepackage{flushend}
\usepackage{amssymb}
\usepackage{amsmath}
\usepackage{breqn}
\usepackage{graphicx}
\usepackage[export]{adjustbox}[2011/08/13]
\usepackage{color}
\usepackage{booktabs}
\usepackage{multirow}
\usepackage{array}
\usepackage{float}
\usepackage{subcaption}
\usepackage{longtable}
\usepackage{bm}
\usepackage{hhline}
\usepackage[ruled]{algorithm2e}
\usepackage{algpseudocode}
\usepackage{eurosym}
\usepackage{url}
\usepackage{cite}
\usepackage{listings}
\usepackage{stackengine} 
\stackMath
\usepackage{changepage}
\usepackage{graphicx}
\usepackage{fancyvrb}

\lstdefinestyle{mystyle}{
	frame=single,
    aboveskip=1mm,
    belowskip=1mm
   }
\usepackage{color} 
\definecolor{mygreen}{RGB}{28,172,0} 
\definecolor{mylilas}{RGB}{170,55,241}
\usepackage{stackengine} 
\stackMath
\usepackage{enumitem}

\usepackage{etoolbox}
\newcommand{\ubold}{\fontseries{b}\selectfont}
\robustify\ubold

\usepackage{multirow}
\usepackage{threeparttable}
\usepackage{siunitx}
\usepackage{multicol}

\makeatletter
\renewcommand{\@seccntformat}[1]{%
  \ifcsname prefix@#1\endcsname
    \csname prefix@#1\endcsname
  \else
    \csname the#1\endcsname\quad
  \fi}
\newcommand\prefix@subsubsection{}
\newcommand\prefix@paragraph{}
\makeatother

\usepackage{hyperref}
\begin{document}
\title{Adversarial Training for a Hybrid Approach to Aspect-Based Sentiment Analysis}
%
%
\author{Ron Hochstenbach\inst{1}, Flavius Frasincar\inst{1}\inst~\orcidID{0000-0002-8031-758X}\and Maria Mihaela Tru\c{s}c\v{a}\inst{2}\inst{(}\Envelope\inst{)}
}
%
%
\institute{
Erasmus University Rotterdam, Burgemeester Oudlaan 50,
	3062 PA Rotterdam, the Netherlands \and Bucharest University of Economic Studies, 010374 Bucharest, Romania \\
	\email{hochstenbach.ron@gmail.com},
    \email{frasincar@ese.eur.nl},
    \email{maria.trusca@csie.ase.ro}}
\maketitle


\begin{abstract}
The increasing popularity of the Web has subsequently increased the abundance of reviews on products and services. Mining these reviews for expressed sentiment is beneficial for both companies and consumers, as quality can be improved based on this information. In this paper, we consider the state-of-the-art HAABSA++ algorithm for aspect-based sentiment analysis tasked with identifying the sentiment expressed towards a given aspect in review sentences. Specifically, we train the neural network part of this algorithm using an adversarial network, a novel machine learning training method where a generator network tries to fool the classifier network by generating highly realistic new samples, as such increasing robustness. This method, as of yet never in its classical form applied to aspect-based sentiment analysis, is found to be able to considerably improve the out-of-sample accuracy of HAABSA++: for the SemEval 2015 dataset, accuracy was increased from 81.7\% to 82.5\%, and for the SemEval 2016 task, accuracy increased from 84.4\% to 87.3\%.
\end{abstract}

\section{Introduction}
    \label{chap: Introduction}

With the ever-increasing popularity of the Web and subsequent increasing abundance of reviews on products or services available, a wealth of interesting information is available to businesses and consumers alike. But due to this increased quantity of reviews available, it becomes more and more infeasible or even impossible to analyse them by hand.

As such, sentiment analysis, 
concerned with the algorithmic analysis of expressed sentiment, can be of great value. In this research, we are interested in Aspect-Based Sentiment Analysis (ABSA), which entails determining the sentiment with respect to a certain aspect\cite{SchoutenSurvey}. Specifically, we perform this task at the sentence level. In the sentence ``The soup was delicious but we sat in a poorly-lit room" for example, a positive sentiment is expressed with regard to the aspect ``Food quality'', but the sentiment towards ``Ambience" is negative.

\cite{wallaart2019hybrid} approaches this problem using a hybrid model, where first an ontology is used to assign a positive or negative sentiment towards an aspect in a rule-based manner. When this method delivers contradicting or no results, an LCR-Rot neural network as described in \cite{zheng2018left}, extended using representation iterations and called LCR-Rot-hop, is used. \cite{trusca2020hybrid} further develops this LCR-Rot-hop model by incorporating contextual word embeddings and hierarchical attention. This new model, called LCR-Rot-hop++, delivers state-of-the-art results.

Research into neural networks is rapidly advancing, and novel techniques are developed continuously. One such technique proposed in \cite{goodfellow2014generative} is Generative Adversarial Network (GAN). Here, two networks are simultaneously trained: a generator tries to generate new input samples, whereas a discriminator tries to discern between real and generated samples. These conflicting objectives converge to a situation where the generator produces samples indiscernible from real data. Furthermore, training a neural network as the discriminator in an adversarial network can increase the robustness of the trained model.

\cite{han2019adversarial} gives an overview of contributions of GANs to the domains of affective computing and sentiment analysis. These mostly fall into the field of image generation, however, which has the benefit that input data is of fixed size. This is not the case in text analysis, as sentences can be of any length. This problem has been worked around in the, so far (to the best of our knowledge), only research where GAN is applied to ABSA in \cite{karimi2020adversarial}. In this paper, not entire new instances are generated, but rather perturbations are made to real data aimed at increasing loss. These are then used to train a BERT Encoder \cite{xu2019bert}, leading to increased accuracy with respect to the baseline model.

This research contributes to the literature in two ways. First, we will develop a method where adversarial samples are fully generated, rather than obtained from perturbing existing samples as in \cite{karimi2020adversarial}. Second, HAABSA++ is a more sophisticated methodology with respect to the specifics of the ABSA task than the BERT Encoder used in \cite{karimi2020adversarial}, making it interesting to investigate whether we can achieve a similar increase in accuracy when applying adversarial to HAABSA++.

This research is structured as follows. First, Sect. \ref{chap: Related Work} gives a brief overview of the related works on ABSA and adversarial training. In Sect. \ref{chap: Data} the datasets used are described. We discuss the HAABSA++ framework in Sect. \ref{chap: Methodology}, as well as how we apply adversarial training to it. Then, Sect. \ref{chap: Results} presents the results of the proposed methods. Last, in Sect. \ref{chap: conclusion} we draw conclusions and give suggestions for further research.

\section{Related Work}\label{chap: Related Work}

\cite{SchoutenSurvey} provides a comprehensive survey on Aspect-Based Sentiment Analysis (ABSA) based on three types of methods: knowledge-based, machine learning, or hybrid approaches. In ABSA, one tries to find the sentiment expressed towards a given or extracted aspect in a sentence or full review (in this research we focus on sentences). Whereas ABSA also comprises tasks like target extraction (i.e., the selection of a target word indicative of a certain aspect) and aspect detection (i.e., the detection of aspects towards which a sentiment is expressed in a text), in this research we follow \cite{trusca2020hybrid} and \cite{wallaart2019hybrid} in focusing on the sentiment classification of sentences in which aspects are explicitly stated in so-called targets. For example, in the sentence ``The food was bland but at least the waiter was allright.", ``food" is a target for the aspect ``Food Quality", whereas ``waiter" serves as a target for the aspect ``Service". Towards the former, a negative aspect is expressed, and towards the latter the sentiment is neutral. 

In \cite{wallaart2019hybrid}, the sentiment classification task of ABSA is addressed by means of a hybrid approach. Here, first a knowledge-based ontology method is used to classify the sentiment towards a target, and a neural network based on \cite{zheng2018left} is used as back-up when this fails. \cite{trusca2020hybrid} extends upon this work by adding two features to the neural network part: contextual word embeddings are used rather than non-contextual ones, and a hierarchical attention mechanism is applied.  Compared to other methods for ABSA, HAABSA++ is found to provide state-of-the-art results when applied to the SemEval 2015 \cite{semeval2015} and SemEval 2016 \cite{semeval2016} datasets.

The field of machine learning is in rapid development, and one particular advancement sparking great interest was proposed in \cite{goodfellow2014generative}. GANs are trained by pitting a generative model against a discriminator. The generator's task is to generate realistic samples when compared to an empirical dataset. At the same time, the discriminator is trained to discern between generated and real samples. These conflicting objectives lead to a minimax game which \cite{goodfellow2014generative} shows to converge to a situation where generated samples are indistinguishable from real samples.

\cite{han2019adversarial} provides an overview of applications of adversarial training to the field of sentiment analysis. Here, the authors note four benefits of adversarial training. First, it allows for generated emotions to feel more natural than those generated through other techniques. Second, it aids in overcoming the sparse availability of labeled sentiment data. Although plenty of images, speech, and text samples exist, only a small fraction of these are labeled and thus suitable for supervised learning. Third, GANs have the potential to learn in a robust manner, reducing the problems occurring when samples are gathered from different contexts. For example, the word ``greasy" related to the domain of restaurant reviews can be understood as positive in a fast-food restaurant, but not in a fine dining establishment. Last, GANs allow for easy quality evaluation of the generated samples. Whereas in prior methods for sample generation, humans needed to assess the quality of the samples, the discriminator built into the GAN framework automates this task. Most of the advancements to GANs listed in \cite{han2019adversarial} focus on generating as realistic as possible samples which are highly relevant for, e.g., image generation, but not as much for the analysis of written reviews.

\cite{Odena} proposes a semi-supervised GAN, where the discriminator simultaneously acts as a classifier. Namely, whereas a regular discriminator is only given the task to discern between `Fake' and `Real' data, in this application the `Real' class is divided into the different classes of the original task. Such a model is commonly referred to as a Categorical Generative Adversarial Network (CatGAN). When applied to the MNIST dataset (where the discriminator discerns between `Fake', `0', `1', ... , `8' and `9'), this provided a significant improvement in performance compared to a non-adversarially trained convolutional neural network. Applications of CatGANs are also found in \cite{springenberg2015unsupervised} and \cite{salimans2016improved}, where again the MNIST dataset is used for testing. In the former, the semi-supervised CatGAN was found to outperform most competing models, only marginally lagging behind the best performing model. Furthermore, CatGAN outperformed competing GAN-based models by great margins, a fact also revealed in the study of \cite{salimans2016improved}. 

An application of CatGAN to ABSA is found only recently in \cite{karimi2020adversarial}, but it is not applied in the traditional sense with an explicit generator and discriminator. Rather, real samples are classified by a BERT Encoder (the network intended to train adversarially). The gradient of the loss is then computed with respect to the initial input data, and based on this the original samples are perturbed. These samples are then again fed to the BERT Encoder, and the total loss is computed based on the combined loss of the perturbed and original samples. Although there is no explicit generator or discriminator here, the BERT Encoder is thus adversarially trained in the sense that samples generated with the explicit aim of being difficult to classify, are fed to it. This methodology outperforms the (non-adversarially trained) BERT-PT model \cite{xu2019bert} when applied to the SemEval 2014 dataset.

\section{Data}
    \label{chap: Data}

For our research, we use data posed for the SemEval 2015 and SemEval 2016 contests. Intended to evaluate methods developed for ABSA, these sets contain English review sentences for restaurants. For each of these sentences, one or multiple aspect categories are denoted together with a negative, neutral or positive sentiment.

\begin{center}
\begin{threeparttable}[h!]
\small
\vspace{-6mm}
\caption{Sentiment frequencies in the used datasets}
\label{tab: sentiment frequencies}
\begin{tabular}{cccccccccc}
\toprule
           & \multicolumn{4}{c}{\textbf{SemEval 2015}}                        &  & \multicolumn{4}{c}{\textbf{SemEval 2016}}                        \\ \cline{2-5} \cline{7-10}
           & \textit{Negative} & \textit{Neutral} & \textit{Positive} & Total &  & \textit{Negative} & \textit{Neutral} & \textit{Positive} & Total \\ \cline{1-5} \cline{7-10}
Train Data & 3.2\%             & 24.4\%           & 72.4\%            & 1278  &  & 26.0\%            & 3.8\%            & 70.2\%            & 1879  \\
Test Data  & 5.3\%             & 41.0\%           & 53.7\%            & 597   &  & 20.8\%            & 4.9\%            & 74.3\%            & 650\\ \bottomrule
\end{tabular}
\end{threeparttable}
\end{center}

\noindent Table \ref{tab: sentiment frequencies} shows the distribution over sentiment polarities in the considered datasets. Here, we note that for both data sets, `Positive' polarities are most abundant. Both data sets have a clear minority class. For SemEval 2015, this is the `Negative' class, whereas for SemEval 2016 this is `Neutral'. For the SemEval 2015 data set, a larger disparity between the relative frequencies among the polarities can be noted between the train and test data compared to SemEval 2016. Furthermore, we did not notice any remarkable discrepancies between the aspect category frequencies in the train and test sets.

We directly apply the datasets as obtained in \cite{wallaart2019hybrid} after preprosessing. First, the authors deleted all sentences where sentiment is implicit, as the applied methodology requires an explicitly mentioned target. For 2015, this amounted to 22.7\% of the train data, and 29.3\% of the test data. For 2016, this was 25.0\% and 24.3\%, respectively\footnote{All values in Table \ref{tab: sentiment frequencies} are excluding implicit targets.}. They then processed the remaining sentences using the NLTK toolkit \cite{bird2009natural}, and, last, tokenized the data and lemmatized the words with the WordNet English lexical database \cite{miller1995wordnet}.


\section{Methodology}
    \label{chap: Methodology}

In this section, the methodology we employ for sentiment classification of restaurant reviews is discussed. First, we will describe the original HAABSA++ algorithm in Sect. \ref{sec: metho-haabsa++}. Then, Sect. \ref{sec: catGan} puts forward the methodology for the CatGAN and elaborates on how to adapt the standard HAABSA++ algorithm to serve as the discriminator in this methodology. Furthermore, this section describes the training procedure.

\subsection{HAABSA++}
\label{sec: metho-haabsa++}

HAABSA++ is a hybrid approach to aspect-based sentiment analysis where, as proposed in \cite{wallaart2019hybrid}, first an ontology is used to determine the sentiment expressed towards a given aspect. If this proves inconclusive, a neural network proposed in \cite{wallaart2019hybrid} and improved upon in \cite{trusca2020hybrid} is used as back-up.

\subsubsection{Ontology.}
\label{subsec: metho-ontology}

The ontology is based upon \cite{schouten2018ontology} and consists of three classes. First, \textit{SentimentMention} contains the expressions of sentiment. It consists of three subclasses, the first of which contains words always expressing the same sentiment value regardless of the aspect, the second containing words always expressing the same sentiment that only adhere to specific aspects, and the third containing words for which the sentiment they express depends on the associated aspect. The \textit{AspectMention} class governs the aspect to which a word adheres, and the \textit{SentimentValue} class labels words as expressing either a positive or negative sentiment. For example, the word `expensive' could fall into the \textit{Price Negative Sentiment} subclass of the \textit{SentimentMention} class, the \textit{Price Mention} subclass of the \textit{AspectMention} class, and the \textit{Negative} subclass of the \textit{SentimentValue} class. Aspect sentiment is then determined by assessing whether in a given sentence, a word falls into the \textit{AspectMention} class of a given aspect, and if so, in what \textit{SentimentValue} subclass it falls. 

The ontology-based method is powerful but might prove inconclusive whenever both negative and positive sentiment are detected for an aspect, or whenever the ontology does not cover the lexicalizations present in a sentence. In these cases, we resort to a back-up neural network approach.

\subsubsection{Multi-Hop LCR-Rot Neural Network with Hierarchical Attention and Contextual Word Embeddings.}
\label{subsec: metho-neural network}

\cite{trusca2020hybrid} further develops the Multi-Hop LCR-Rot-hop neural network used as back-up in \cite{wallaart2019hybrid} into LCR-Rot-hop++ by adding contextual word embeddings and a hierarchical attention mechanism. Although they compare multiple ways of implementing these additions, we only consider the BERT contextual word embeddings \cite{BERT} as these are found to perform best out-of-sample in our datasets. Furthermore, we use the last method out of the four methods for hierarchical attention introduced in \cite{trusca2020hybrid}, as it performs best on average. According to the fourth method, the last layer of attention is applied repetitively and separately for contexts and targets.

First, the words in a sentence are turned into word embeddings using the contextual BERT model. Here, first token vectors that are unique for each word, position embeddings showing the word's location in the sentence, and segment embeddings discerning between multiple sentences present, are averaged. The newly computed embeddings feed into a Transformer Encoder. The entire network is pre-trained on the Masked Language Model and Next Sentence Prediction tasks. This method ensures that a word's embedding depends on how it is used in the sentence. For example, the word `light' could refer to an object's mass or to the visible light coming from the sun or a lamp, depending on the context.

Then, the embedded sentences are split into three parts: Left (with respect to the target), Target, and Right (with respect to the target). Each of these feeds a bi-directional LSTM layer. These outputs are then used to create four attention vectors: a target2context and a context2target for both the left and right contexts. This is done using a two-step rotary attention mechanism, where in each iteration the intermediate context and target vectors are weighted using an attention score computed at the sentence level. This weighting is done separately for the two target vectors and the two context vectors. The four resulting vectors are then concatenated and subjected to another attention layer, which computes attention weights separately for contexts and targets. This process is repeated for a specified number of hops, the final output of which is used as input for a multi-layer perceptron determining the final sentiment classification.

\subsection{Classifying Generative Adversarial Network}
\label{sec: catGan}

As described before, a promising advancement in the field of deep learning is the Generative Adversarial Network. First proposed in \cite{goodfellow2014generative}, a GAN generates new data samples by simultaneously training a generative and a discriminative model. While the generative model produces new samples based on a random input vector, the discriminative model tries to discern between the generated samples and real data.

This section first describes how to adapt the neural network in HAABSA++ so as to constitute both a classifier and a discriminator (i.e., a CatGAN). This added objective should increase the robustness of the model, as it is forced to more clearly recognize what input characteristics determine the class to which an instance belongs. This can be made more apparent through an example. Imagine a child tasked with labeling images of vehicles as either `plane', `train', or `car'. It might then be fairly easy for the child to recognize cars, as only they have tires, but such a strategy will lead him to wrongly characterize planes on the ground (with their landing gear out) as cars. If, however, every now and then an image of a bicycle is given to the child, this challenges him to notice that it is not the characteristic ``having tires" by which cars can be recognized, but rather ``having four tires". This realization will in turn lead him to not label grounded planes as cars anymore, as they have more than four tires.

In the first subsection, we describe the model, and the second subsection gives its implementation details. Then, we present the training procedure and discuss convergence in GAN training. The last subsection describes the procedure for hyperparameter optimization.

\subsubsection{Model Formulation.}
\label{subseb: model formulaion}

\cite{salimans2016improved} shows how the loss function of the regular GAN model can be adapted so the discriminator simultaneously serves as a classifier, as such constituting a Categorical GAN (CatGAN). Instead of discriminating between real or fake data, the discriminator now discriminates between K+1 classes. In our case, K=3 and the first three classes are `Negative', `Neutral' and `Positive'. The K+1$^{\text{th}}$ class is `Fake'. This yields the loss function:

\begin{equation}
\label{eq: implicit loss function}
\begin{aligned}
    L_{G,D} = &-\mathbb{E}_{\Vec{x},y \sim p_{\text{data}}(\Vec{x},y) }\text{log}[p_{\text{model}}(y|\Vec{x},y<\text{K}+1)] + \lambda(||\Theta_G||^2 +  ||\Theta_D||^2) \\
    &- \mathbb{E}_{\Vec{z}\sim P(\vec{z})}\text{log}[p_{\text{model}}(y=\text{K}+1|\Vec{z})],
\end{aligned}
\end{equation}

\noindent resulting in the following optimization problem to solve:

\begin{equation}
\label{eq: minimax problem}
    \max_G \min_{D} L_{G,D}.
\end{equation}

\noindent In Equation \ref{eq: implicit loss function}, the first two terms correspond to the loss term as posed in \cite{wallaart2019hybrid}\footnote{Note that $\lambda||\Theta_G||^2$ and $\lambda||\Theta_D||^2$ correspond to L$_2$-regularization terms and are not included in \cite{salimans2016improved}.}. The last term corresponds to the the samples generated by the generator, which are labeled to the new, `Fake' class ($y = K+1 = 4$). Furthermore, $y=1$ means `Negative', $y=2$ is `Neutral' and $y=3$ is `Positive'. This loss function can be split into loss functions for the generator and discriminator, respectively, as follows:

\begin{subequations}
   \begin{align}
     L_{G} = &- \mathbb{E}_{\Vec{z}\sim P(\vec{z})}\text{log}[p_{\text{model}}(y=\text{K}+1|\Vec{z})] + \lambda||\Theta_G||^2\\
     L_{D} = &-\mathbb{E}_{\Vec{x},y \sim p_{\text{data}}(\Vec{x},y) }\text{log}[p_{\text{model}}(y|\Vec{x},y<\text{K}+1)] + \lambda  ||\Theta_D||^2 \\
    &- \mathbb{E}_{\Vec{z}\sim P(\vec{z})}\text{log}[p_{\text{model}}(y=\text{K}+1|\Vec{z})]\nonumber.
   \end{align}
\end{subequations}

\noindent Note here that the last term of Equation \ref{eq: implicit loss function} is included in both the generator and discriminator loss functions, as fake samples are first processed by the generator and then labeled by the discriminator. Furthermore, one could rewrite $\text{log}[p_{\text{model}}(y=\text{K}+1|\Vec{z})]$ as $\text{log}[1-D(G(\vec{z}))]$ and $\text{log}[p_{\text{model}}(y|\Vec{x},y<\text{K}+1)]$ as $\text{log}[D_y(\vec{x})]$, where $D(\vec{x})=\sum\limits_{y=1,2,3}D_y(\vec{x})$ represents the probability that the argument $\vec{x}$ stems from the real data. $D_y(\vec{x})$ in turn represents the probability that the argument stems from the real data and has label $y$.

Written explicitly as in \cite{wallaart2019hybrid}, Equation \ref{eq: implicit loss function} becomes:

\begin{equation}
\label{eq: explicit loss function}
    L_{G,D} = -\sum_{j \in \mathbb{J}}  \vec{y}_j \times \text{log}(\hat{p}_j) - \sum_{i \in \mathbb{I}} \vec{y}_i \times \text{log}(\hat{p}_i)  + \lambda(||\Theta_G||^2 + ||\Theta_D||^2).
\end{equation}

\noindent Here, $j \in \mathbb{J}$ is the batch of real data samples. $\vec{y}_j$ is the real sentiment of the $j^{th}$ sample written in vector form, i.e., when $y_j=1$, $\vec{y}_j=[1,0,0,0]$, when $y_j=2$, $\vec{y}_j=[0,1,0,0]$ etc. $\hat{p}_j$ is the predicted sentiment of the $j^{th}$ sample. $i \in \mathbb{I}$ are the generated samples, and $\hat{p}_i$ is the predicted sentiment for the $i^{th}$ generated sample. Note furthermore that $\vec{y}_i$ is always equal to $[0,0,0,1]$. $\Theta_G$ and $\Theta_D$ contain the parameters for the generator and discriminator, respectively.

A problem occurs, however, when a generator is tasked with forming sentences. Constituted by a neural network, the generator can only output vectors and thus sentences of fixed length. This poses no problem in classical applications of GANs, as for example images generated always have the same resolution and thus output length. Sentences have variable length, however, and the discriminator can therefore possibly distinguish between generated and real samples based on their length. As such, we here only generate the four LCR-Rot-hop++ representation vectors of length $2d$ each, where $d$ is the dimensionality of the embedding. Then, we use the final MultiLayer Perceptron (MLP) of the LCR-Rot-hop++ network simultaneously as classifier and discriminator. The only modifications necessary are then to adapt the loss function to Equation \ref{eq: explicit loss function}, and let the MLP output a vector of probabilities of length $K+1$ rather than $K$.

\begin{figure}[h!]
\begin{center}
\includegraphics[width=0.8\textwidth]{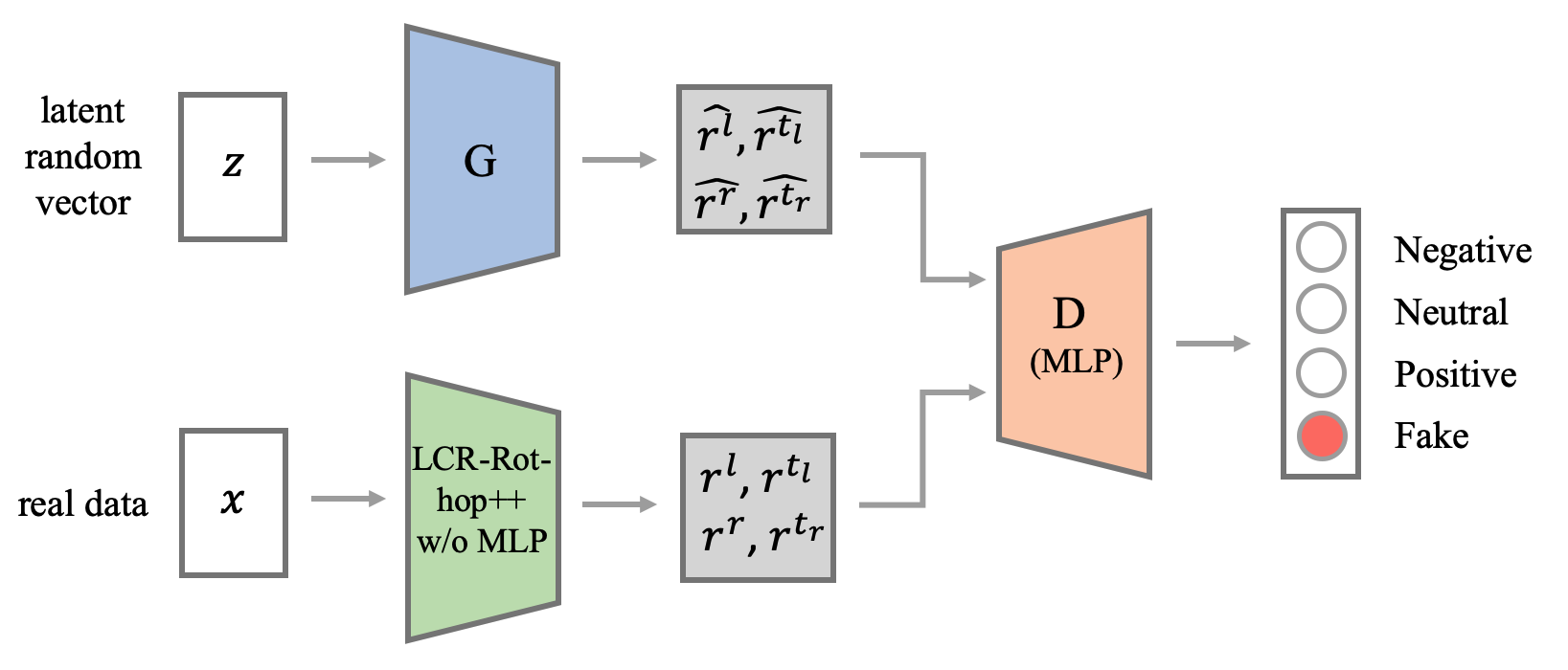}%
\end{center}
\caption{HAABSA* structure}
    \label{Figure: HAABSA* Schematic}
\vspace{-6mm}
\end{figure}

The generator is constituted by a fully connected, 4-layer MLP. It uses an $r$-dimensional random vector as input, the first hidden layer will have $2 \times d$ neurons and the second hidden layer will have $6 \times d$ neurons. It then outputs a vector of length $8d$ (according to the LCR-Rot-hop++ model, the four vectors of length 2d are concatenated, resulting a vector of length 8d). This choice of architecture is the result of a trade-off between model complexity and speed of training. Then, the discriminator is constituted by the final MLP layer of the Multi-Hop LCR-Rot++ network in HAABSA++. A schematic of this new algorithm, called HAABSA*, is given in Figure \ref{Figure: HAABSA* Schematic}.

\subsubsection{Implementation Details.}
\label{subsec: implementation details}
Following \cite{wallaart2019hybrid} and \cite{trusca2020hybrid}, we run the optimization algorithm for 200 iterations. For each iteration, a batch of $m=20$ real samples, as well as 20 fake samples, is selected. We initialize all weights randomly following a U(-0.01,0.01) distribution, and all biases are initialized as zero. To ensure all parameters are included in the regularization terms, $\Theta_D$ includes both the parameters in the final MLP layer and the parameters in the LCR-Rot-hop++ network w/o MLP. $\Theta_G$ includes the parameters of the generator. The random input for the generator follows a U(0,1) distribution and is of dimension $r=100$. As in \cite{trusca2020hybrid}, the dimensionality of the word embedding $d=768$. Last, it is important to note that that, like \cite{wallaart2019hybrid} and \cite{trusca2020hybrid}, we always train the adversarial network on the full training set. This may result in bias, as for the test set, only samples not classified by the ontology are classified by the neural network. However, the increased performance due to the larger training set size outweighs this downside.
Our methods are implemented in \verb+Python+  using the \verb+TensorFlow+  platform. All code can be found at \url{https://github.com/RonHochstenbach/HAABSAStar}.

\subsubsection{Training Procedure and GAN Convergence.}
\label{subsec: Training procedure and GAN convergence}

The training procedure for HAABSA*, based on the one put forward in \cite{goodfellow2014generative}, is shown in Algorithm \ref{Algorithm: HAABSA*}. In our implementation, the discriminator is updated every iteration, whereas the generator is updated only every $k^{th}$ iteration \cite{goodfellow2014generative}.

\begin{algorithm}[h!]
\small
\SetAlgoLined
 \For{number of training iterations}{
  \eIf{iteration number divisible by k}{
   \begin{itemize}
	\item Sample minibatch of $m$ noise samples $\{\Vec{z}^{(1)},\ldots,\Vec{z}^{(m)}\}$ from noise
	\\prior $p_{\text{noise}}(\Vec{z})$
	\item Sample minibatch of $m$ examples with their associated labels $\{(\Vec{x}^{(1)},y^{(1)}),\ldots,(\Vec{x}^{(m)},y^{(m)})\}$ from dataset $\mathcal{X}$
	\item Update the generator by ascending  the stochastic gradient of the
	\\loss function with respect to $\Theta_G$:
	
	\begin{equation*}
	    \nabla_ {\Theta_G}-\frac{1}{m}\sum_{i=1}^m[\text{log}(1-D(G(\vec{z}^i)))].
	\end{equation*}
	
	\item Update the discriminator by descending the stochastic gradient
	\\of the loss function with respect to $\Theta_D$:
	
	\begin{equation*}
	    \nabla_ {\Theta_D}-\frac{1}{m}\sum_{i=1}^m[\text{log}(D_y(\vec{x}^i))+\text{log}(1-D(G(\vec{z}^i)))].
	\end{equation*}
	\end{itemize}
   }{
   \begin{itemize}
				\item Sample minibatch of $m$ noise samples $\{\Vec{z}^{(1)},\ldots,\Vec{z}^{(m)}\}$ from noise
				\\prior $p_{\text{noise}}(\Vec{z})$
				\item Sample minibatch of $m$ examples with their associated labels $\{(\Vec{x}^{(1)},y^{(1)}),\ldots,(\Vec{x}^{(m)},y^{(m)})\}$ from dataset $\mathcal{X}$
				\item Update the discriminator by descending the stochastic gradient
				\\of the loss function with respect to $\Theta_D$
				\end{itemize}
  }
 }
 \caption{HAABSA* training procedure}
 \label{Algorithm: HAABSA*}
\end{algorithm}
 
\cite{goodfellow2014generative} proves the minimax game given in Equation \ref{eq: minimax problem} has a theoretical optimum under two conditions. First of all, the generator and discriminator need enough capacity to, respectively, generate realistic samples and discern them effectively. No formal method to ascertain sufficient capacity is given, however. Second, the discriminator should be allowed to reach its optimum given an instance of the generator, before the generator is updated again. In Algorithm \ref{Algorithm: HAABSA*}, this can be (amongst other later to be discussed options) controlled by the factor $k$.

However, in \cite{goodfellowNonConverge}, it is explained that this only holds true for the convex case. Due to the minimax nature of the game in which a saddle-point is the optimum, convergence is not guaranteed. Namely, the updating procedures for both networks have conflicting aims, and updates with respect to, for example, $\Theta_G$ might counteract gains made in prior updates to $\Theta_D$. This means it could also occur that the generator, instead of increasing the loss function, makes the bound in which the discriminator can update looser. This results in the algorithm oscillating instead of converging.
\cite{salimans2016improved} describes this problem as well, and further notes that when either the generator or discriminator becomes too good at its task, this prevents the other from learning further. For example, when the discriminator perfectly identifies all fake samples, the generator has no reference for how to change its tactic in order to make samples capable of being misclassified as real by the discriminator. 

\subsubsection{Hyperparameter Optimization.}
\label{subsec: hyperparam optim}

\cite{salimans2016improved} mentions several intricate possible techniques that could increase the probability of convergence, but careful selection of hyperparameters could also at least prevent divergence. \cite{trusca2020hybrid} considers multiple hyperparameters, two of which could aid in preventing divergence.

First, the learning rate of the employed momentum optimizer governs the pace at which descent takes place in the gradient direction. Whereas too small values might prevent reaching of the optimum in the given number of iterations, smaller values prevent one of the networks from becoming too good too quickly for the other to keep up. Second, the momentum term governs the extent to which gradient values of previous iterations are considered \cite{sutskever}. Lower values for this might, in a similar manner as with the learning rate, prevent divergence.

\cite{trusca2020hybrid} also considers the L$_2$-regularization term and the dropout probability (which has been applied to all hidden layers of the network to prevent overfitting) as hyperparameters. Since, however, the optimal value for these was not found to differ much compared to \cite{wallaart2019hybrid}, and they do not directly affect the convergence of the minimax game, we leave them at the values found in \cite{trusca2020hybrid} (i.e. 0.0001 and 0.3 for the L$_2$-regularization term and keep probability, respectively).

Preliminary trial runs showed that higher values for the learning rate and momentum term led the generator to cause divergence. However, lower values impeded the discriminator from learning. As such, we use different values for the learning rate and momentum term for the generator and discriminator, writing the learning rate and momentum term for the generator as a function of those terms for the discriminator as follows: $lr_{gen} = \mu_{lr} \times lr_{dis}$ and $mom_{gen} = \mu_{mom} \times mom_{dis}$. Here, $\mu_{lr}$ and $\mu_{mom}$ are multiplier terms for the learning rate and momentum, respectively. We will then consider $lr_{dis}$, $mom_{dis}$, $\mu_{lr}$ and $\mu_{mom}$ as hyperparameters for optimization.

Last, we treat the value of $k$ as a hyperparameter. As this value determines how much more often the discriminator is updated then the generator, it can be of great importance in ensuring one of the networks does not learn too much more quickly than the other.

For the hyperparameter optimization, we use a Tree-structured Parzen Estimator (TPE) approach \cite{bergstra2011}. This method has a higher convergence speed than other methods like grid search. It optimizes by, for each run, selecting hyperparameters from the options given by maximizing the ratio of fit values of the best set of hyperparameters to the fit values of the previously unconsidered options for hyperparameters.

We run the hyperparameter optimization for the 2015 and 2016 datasets separately. The considered options for the learning rate and momentum term of the discriminator are based upon the found values in \cite{trusca2020hybrid}. To ensure a manageable runtime, we limit the algorithm to 20 runs. The algorithm is run on datasets obtained by randomly splitting the training set of the given year into 80\% train and 20\% test data.


\section{Results}
    \label{chap: Results}

In this section, we discuss the results obtained from the adversarial training of HAABSA++. First, we provide a brief discussion of the results of hyperparameter optimization in Sect. \ref{sec: Results hyperpar optim}. Then, Sect. \ref{sec: performance analysis} puts forward the obtained accuracies using HAABSA*. Furthermore, we compare these with the accuracies of HAABSA++ and other competing models for ABSA.

\subsection{Results Hyperparameter Optimization}
\label{sec: Results hyperpar optim}


To decrease the risk of divergence, we include different values for the learning rate and momentum term of the generator and discriminator. Whereas trial runs with equal learning rates or momentum terms for the generator and discriminator diverged on numerous occasions, for different learning rates and momentum terms the hyperparameter optimization only one out of the total of 40 runs (20 for each dataset) diverged. The results of the hyperparameter optimization are shown in Table \ref{tab: hyperparam}.

\begin{center}
\begin{threeparttable}[h!]
\small
\caption{Hyperparameter optimization results }
\label{tab: hyperparam}
\begin{tabular}{llcc}
\toprule
\multicolumn{1}{c}{\textbf{Hyperparameter}} & \multicolumn{1}{c}{\textbf{Considered Values}} & \multicolumn{2}{l}{\textbf{Optimal Value}} \\
\multicolumn{1}{c}{}                        & \multicolumn{1}{c}{}                           & \textbf{2015}        & \textbf{2016}       \\ \hline
$lr_{dis}$                                  & [0.007, 0.01, 0.02, 0.03, 0.05, 0.09]          & 0.02                 & 0.03                \\
$mom_{dis}$                                 & [0.7, 0.8, 0.9]                                & 0.9                  & 0.7                 \\
$\mu_{lr}$                                  & [0.1, 0.15, 0.2, 0.4]                          & 0.1                  & 0.15                \\
$\mu_{mom}$                                 & [0.4, 0.6, 0.8, 1.6]                           & 0.4                  & 0.6                 \\
$k$                                         & [3, 4, 5]                                      & 3                    & 3                   \\ \bottomrule
\end{tabular}
\end{threeparttable}
\vspace{-3mm}
\end{center}

\begin{center}
\begin{threeparttable}[h!]
\small
\caption{Accuracies of HAABSA++ and HAABSA*}
\label{table: accuracies}
\begin{tabular}{clccccc}
\toprule
\multicolumn{1}{l}{}                    &                   & \multicolumn{2}{c}{\textbf{SemEval 2015}}   &  & \multicolumn{2}{c}{\textbf{SemEval 2016}}   \\ \cline{3-4} \cline{6-7}
\multicolumn{1}{l}{}                    &                   & \textit{in-sample} & \textit{out-of-sample} &  & \textit{in-sample} & \textit{out-of-sample} \\ \hline
\multirow{2}{*}{\textbf{w ontology}}   & HAABSA++ & 88.8\%             & 81.7\%                 &  & 91.0\%             & 84.4\%                 \\
                                        & HAABSA*  & 89.7\%             & \textbf{82.5\%}                 &  & 91.5\%             & 87.3\%                 \\
\multirow{2}{*}{\textbf{w/o ontology}} & HAABSA++ & 94.9\%             & 80.7\%                 &  & 95.1\%             & 80.6\%                 \\
                                        & HAABSA*  & \textbf{96.6\%}             & 82.2\%                 &  & \textbf{96.2\%}             & \textbf{88.2\%}                 \\ \bottomrule
\end{tabular}
\begin{tablenotes}[flushleft]
\small
\item 
Note that the accuracies found for HAABSA++ differ slightly from those reported in \cite{trusca2020hybrid}. This might be due to the stochastic nature of batch selection during training, which was not seeded in the implementation.
Best result per dataset is shown in bold.
\end{tablenotes}
\end{threeparttable}
\vspace{-2mm}
\end{center}

\subsection{Performance Analysis}
\label{sec: performance analysis}

We ran our algorithm using the hyperparameters with the values discussed in the previous section. The results are displayed in Table \ref{table: accuracies}. A few considerations must be made when interpreting these results. HAABSA++ was developed as a hybrid approach, since the ontology accuracy for the cases that it was able to predict was found to be higher than the neural network accuracy out-of-sample (82.8\% versus 81.7\% and 86.8\% versus 84.4\% for the SemEval 2015 and 2016, respectively). To get a better understanding of how the neural network parts of HAABSA++ and HAABSA* perform, we also report the accuracies without ontology. Furthermore, training the HAABSA* assumes the classification of data in four classes, whereas HAABSA++ only has three classes to discern between. This hinders direct comparison between the in-sample accuracies of the two models. However, the evaluation of the testing accuracy is more straightforward as we consider only three sentiment options `Negative', `Neutral' and `Positive' for both HAABSA ++ and HAABSA*.

For our comparison between HAABSA++ and HAABSA*, we will thus focus on out-of-sample accuracy. Here, we find HAABSA* to outperform HAABSA++ by a considerable margin. Specifically, when only comparing the neural network parts of both algorithms (i.e., accuracy without ontology), a large increase in accuracy of 7.6 percentage points is found to be achieved by the adversarial training procedure for 2016. For 2015, a smaller but still substantial increase in accuracy of 1.5 percentage points is found.

It is interesting to note that for the SemEval 2016 task, a higher accuracy is obtained when not using the ontology compared to when the ontology is used. For SemEval 2015, using the ontology does yield an increase in accuracy still, but with 0.3 percentage points, the margin is small. After further development of the adversarial training method for the LCR-Rot-hop++ method for ABSA, it might be worthwhile to opt out of using the ontology (and thus, the hybrid approach), instead only using the neural network part.

\begin{center}
\begin{threeparttable}[b]
\small
\caption{Comparison with competing models}
\label{tab: comparison models}
\begin{tabular}{lcllc}
\toprule
\multicolumn{2}{c}{\textbf{SemEval 2015}}        &  & \multicolumn{2}{c}{\textbf{SemEval 2016}}        \\ \cline{1-2} \cline{4-5}
HAABSA* (w/o ontology) & \textbf{82.2\%}                  &  & HAABSA* (w/o ontology) & \textbf{88.2\%}                  \\
HAABSA++ \cite{trusca2020hybrid}             & 81.7\% &  & XRCE \cite{pontiki-etal-2014-semeval}                  & 88.1\% \\
LSTM+SynATT+TarRep \cite{He-effectiveattentionmodeling}      & 81.7\% &  & HAABSA++ \cite{trusca2020hybrid}               & 87.0\% \\
PRET+MULT \cite{pretMult}               & 81.3\% &  & BBLSTM-SL \cite{BBLSTM}             & 85.8\% \\
BBLSTM-SL \cite{BBLSTM}              & 81.2\% &  & PRET+MULT \cite{pretMult}             & 85.6\% \\\bottomrule
\end{tabular}
\begin{tablenotes}[flushleft]
\small
\item
Values shown are out-of-sample accuracies as reported in the respective research.
Best result per dataset is shown in bold.
\end{tablenotes}
\end{threeparttable}
\end{center}

As can be seen in Table \ref{tab: comparison models}, this increase in accuracy leads to HAABSA* outperforming LSTM+SynATT+TarRep (81.7\%) for the SemEval 2015 task. Furthermore, when the ontology is not used, HAABSA* outperforms XRCE (88.1\%) for the SemEval 2016 task. As such, it performs best among those models considered for the SemEval 2015 and 2016 tasks in \cite{trusca2020hybrid}.

It is also of interest to compare our obtained increase in accuracy from adversarial training with the results of \cite{karimi2020adversarial}.
Although they differ from the original implementation of GANs, instead opting for adversarial perturbations, this is, to the best of our knowledge, the only other research in which a form of adversarial training is applied to ABSA. Whereas direct comparison is not possible since they consider different datasets and a different benchmark model, we note that \cite{karimi2020adversarial} reports increases in accuracy with a maximum of 1 percentage point for all considered datasets and tasks with respect to the non-adversarially trained benchmark model. We, however, obtain an increase of 7.6 percentage points by training our neural network adversarially on the SemEval 2016 task, while an increase of 1.5 percentage points is found for the 2015 task. This could indicate that the classic implementation of GANs
\cite{goodfellow2014generative} with explicit generator and discriminator delivers better results than the approach with adversarial perturbations.


\section{Conclusion}
    \label{chap: conclusion}

In this work, we applied adversarial training to the neural network part of the method devised for ABSA in \cite{trusca2020hybrid}. Although convergence might prove troublesome for adversarial networks, we found enabling differing learning rates and momentum terms for the generator and discriminator to reduce the problem of divergence. As such, we achieved an increase overall performance from 81.7\% to 82.5\% and from 84.4\% to 87.3\% for the SemEval 2015 and 2016 tasks, respectively. When only using the neural network part, adversarial training increased accuracy from 80.6\% to 88.2\% for the 2016 task, thus outperforming the hybrid approach with ontology. For the 2015 task, accuracy increased from 80.7\% to 82.2\%.
These increases in accuracy are considerably higher than those found in \cite{karimi2020adversarial}: the, so far, only other application of adversarial training to ABSA, where the authors opted for adversarial perturbations rather than the classical implementation with explicit generator and discriminator.

In future work, we would like to investigate other methods ensuring stability in GAN training, such as feature matching 
\cite{featurematching1},
, minibatch discrimination \cite{salimans2016improved}, and historical averaging \cite{salimans2016improved}.  Furthermore, we wish to explore whether the generator network can be improved by producing full sentences rather than attention vectors, and experiment with other architectures for the generator. Last, we plan a more direct comparison to adversarial perturbation method of \cite{karimi2020adversarial} by using the same datasets and benchmark model.

\bibliographystyle{splncs04}
\bibliography{references}
\end{document}